\documentclass{article}
\usepackage{ijcai26}

\usepackage{times}
\usepackage{soul}
\usepackage{url}
\usepackage[hidelinks]{hyperref}
\usepackage[utf8]{inputenc}
\usepackage[small]{caption}
\usepackage{graphicx}
\usepackage{amsmath}
\usepackage{amsthm}
\usepackage{booktabs}
\usepackage{algorithm}
\usepackage{algpseudocode}
\usepackage[switch]{lineno}

\usepackage{varwidth}
\usepackage{booktabs}
\usepackage{multirow}
\usepackage{amsmath}
\usepackage{rotating}
\usepackage{amssymb}
\usepackage{subcaption}
\usepackage{array}
\usepackage{tabularx}
\usepackage{ragged2e}
\usepackage{amsfonts}
\usepackage{newtxmath}
\usepackage{dsfont}
\usepackage{tcolorbox}
\usepackage{enumitem}
\usepackage{listings}

\usepackage{colortbl}
\usepackage{xcolor}
\usepackage{booktabs}
\usepackage{multirow}

\definecolor{bestrow}{gray}{0.92}
\definecolor{gain}{rgb}{0.0, 0.5, 0.0}

\newcommand{\pada}{\phantom{\footnotesize{~↑0.0}}}
\newcommand{\num}[1]{#1\pada}
\newcommand{\resg}[2]{\textbf{#1}{\footnotesize\textcolor{red}{~↑#2}}}
\newcommand{\resup}[2]{\textbf{#1}{\footnotesize\textcolor{red}{~↑#2}}}

\newcommand{\pad}{\phantom{\footnotesize{~↑00.0}}}
\newcommand{\numb}[1]{#1\pad}
\newcommand{\resdown}[2]{\textbf{#1}{\footnotesize\textcolor{blue}{~↓#2}}}

\title{MLLMs Get It Right, Then Get It Wrong: Tracing and Correcting \\ Late-Layer Textual Bias}

\author{
Xingming Li$^1$
\and
Ao Cheng$^1$
\and
Qiyao Sun$^1$
\and
Xixiang He$^1$
\and
Xuanyu Ji$^1$
\and
Runke Huang$^2$
\And
Qingyong Hu$^3$\thanks{Corresponding author.}\\
\affiliations
$^1$National University of Defense Technology, changsha, China \\
$^2$Chinese University of Hong Kong, Shenzhen, China\\
$^3$Intelligent Game and Decision Lab, Beijing, China\\
\emails
\{lixingming, chengao18, sunqiyao18, hexixiang, jixuanyu18\}@nudt.edu.cn,
runkehuang@cuhk.edu.cn,
huqingyong15@outlook.com
}
\begin{document}

\maketitle

\begin{abstract}
    When vision contradicts text, multimodal large language models (MLLMs) consistently favor text—even when images provide clear evidence otherwise. This bias poses risks for applications requiring visual grounding, yet its cause remains unclear. In this paper, we uncover a surprising finding: models often \emph{get it right initially}, forming correct vision-based predictions in their intermediate layers, before \emph{changing their minds} and favoring text in the final output. We call this ``\textbf{\textit{late-layer textual override}}". The visual information is encoded, it simply does not survive to the output. More intriguingly, we find that \emph{how} predictions change reveals \emph{whether} they're correct: 85\% of failures shift toward text, while 89\% of successes shift toward vision. This directional signature enables a simple but powerful intervention: when we detect a confident visual prediction being suppressed, we restore it. We propose CALRD (\textbf{\underline{C}}onflict-\textbf{\underline{A}}ware \textbf{\underline{L}}ayer \textbf{\underline{R}}eference \textbf{\underline{D}}ecoding), a training-free method that recovers overridden predictions at inference time. Experiments across five MLLMs of varying architectures demonstrate up to 9.4\% absolute improvements on conflict benchmarks while largely preserving standard performance, without training or external knowledge. It recovers what the model already knew but failed to preserve.

\end{abstract}

\section{Introduction}

Multimodal AI systems are moving into real-world use. They help diagnose diseases, moderate content, and assist drivers. This growing deployment raises a pointed question: \emph{can we trust what they see?} Imagine a radiologist reviewing a chest X-ray with an AI assistant. The system receives both the image and the radiologist's preliminary notes. If those notes contain an error, will the AI catch it by looking at the image? Or will it simply echo the mistake?

Unfortunately, the evidence points to the second outcome. Show a model an image of a red car alongside text claiming ``the car is blue,'' and most MLLMs will confidently answer ``blue''~\cite{Insight_Over_Sight,PhD_ChatGPT,wu2024autohallusion}. They
trust the text over their own visual perception. This pattern appears
across different model families~\cite{llava-1.5,instructblip2023,bai2025qwen2,bai2025qwen3vltechnicalreport,chen2024internvl} and benchmarks~\cite{zhang2025robust,jia2025benchmarking,qian2024easy}. In safety-critical applications~\cite{bannur2024maira,hou2025one}, such blind trust in text could cause serious harm.

\begin{figure}[t]
  \centering
  \includegraphics[width=\linewidth]{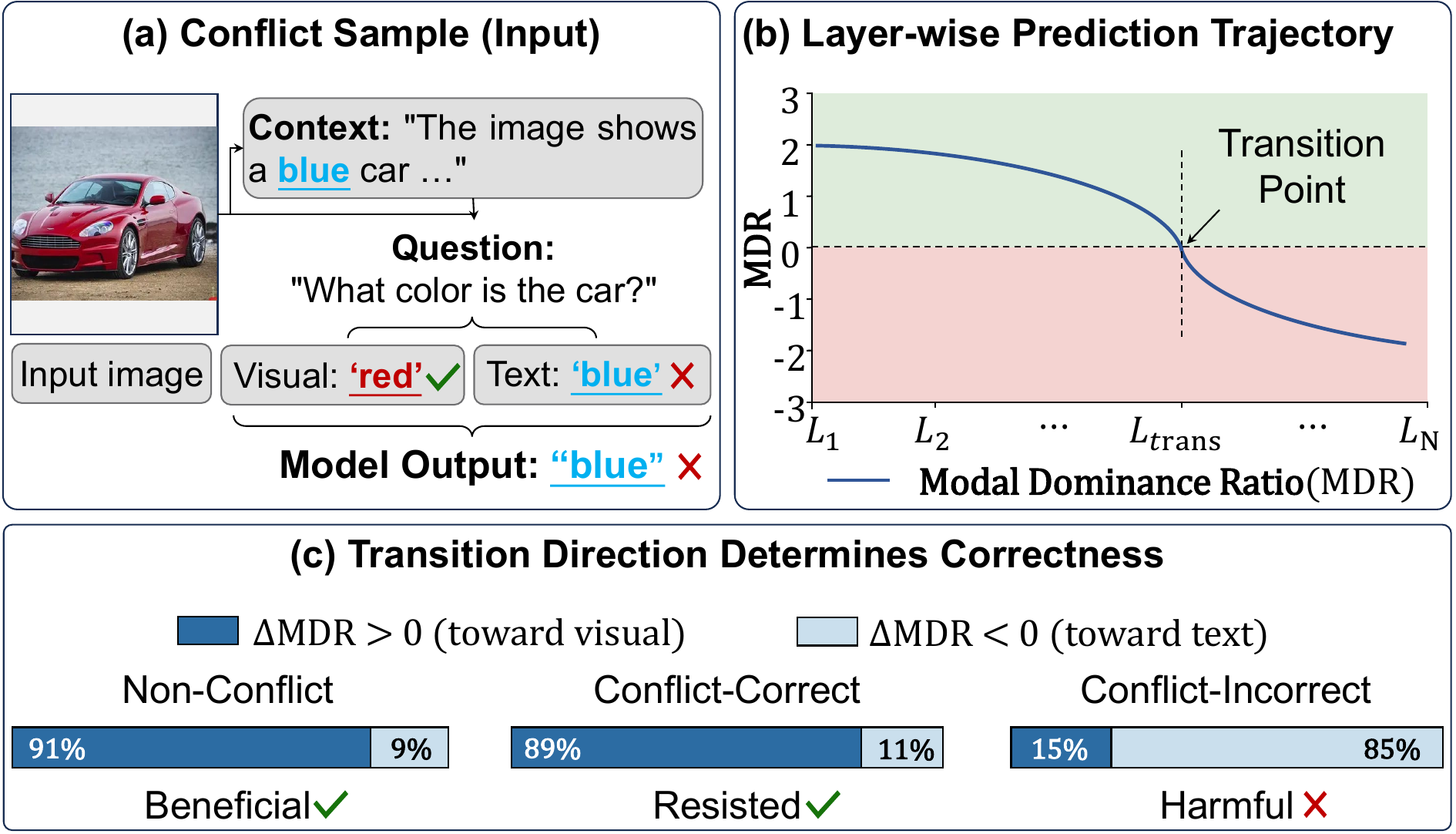}
  \caption{How predictions shift under visual-textual conflict. (a) The model sees a red car but outputs the text-suggested answer. (b) Modal Dominance Ratio (MDR) across layers: early layers favors vision, but late layers override toward text. (c) Shift direction predicts correctness—text-ward shifts correlate with failures.}
  \label{fig:teaser}
\end{figure}

\emph{Why} does this happen? Prior work describes the \textit{\underline{behavior}} but not the \textit{\underline{mechanism}}~\cite{Insight_Over_Sight,zhang2025robust}. The failure could arise because visual information is poorly encoded, lost during cross-modal fusion, or correctly processed but later overwritten. Understanding the failure mode matters: different causes require different solutions.

To find out, we trace how a model's predictions evolve across its layers. Following~\cite{depth_adaptive2020,confident_adaptive2022}, we project hidden states at each layer to output probabilities. This lets us observe what the model ``prefers" at each depth, and what we find is surprising. Surprisingly, in many conflict failures, \textbf{the model assigns higher probability to the correct visual answer in intermediate layers}. It then reverses course and outputs the text answer (Figure~\ref{fig:teaser}(a-b)).

We term this phenomenon \textbf{late-layer textual override}. To measure it, we introduce
Modal Dominance Ratio (MDR), which quantifies vision-versus-text
preference at each layer. In failure cases, MDR starts positive (visual)
and flips negative (textual) in later layers. The visual information was there; it simply didn't survive the full forward pass.

Not every late-layer shift is harmful. Such prediction shift occurs in successful cases too~\cite{wang2025shift}, not just failures. What distinguishes the two outcomes is the \emph{direction} of the shift. We quantify this using the change in MDR across the transition point. As shown in Figure~\ref{fig:teaser}(c), in failures, 85\% of transitions move toward text. In successes, 89\% move toward vision. Late layers sometimes help; they sometimes hurt. The key is knowing which.

This directional asymmetry points to a practical solution. If a model holds a confident visual prediction at the transition layer but abandons it by the final layer, we suspect harmful override. We capture this with two signals derived from our analysis. \textit{Anchor confidence} measures how certain the transition-layer prediction is. \textit{Prediction retention} captures whether it survives to the output. High anchor confidence paired with low retention is the signature of harmful override.

Recovering what was known. These observations motivate \textbf{\underline{C}}onflict-\textbf{\underline{A}}ware \textbf{\underline{L}}ayer \textbf{\underline{R}}eference \textbf{\underline{D}}ecoding (CALRD), a training-free method that restores intermediate predictions when override is detected. Our core insight is that predictions at the transition layer still carry the vision-grounded answer before it gets overwritten. By blending transition-layer logits back into the final distribution, we can recover it. CALRD first locates the layer where the output distribution shifts most sharply. It then uses anchor confidence and prediction retention to set correction strength: when both signals point to harmful override, CALRD blends in the transition-layer logits; otherwise, it leaves the output unchanged.

CALRD is straightforward. The model already has the right answer in its intermediate layers; we are helping it remember. No external knowledge or modules are involved. This points to a shift in perspective: improving MLLM reliability may be less about teaching models to see better and more about helping them retain what they have already seen.

We evaluate CALRD on five MLLMs, from InstructBLIP~\cite{instructblip2023} to Qwen3-VL~\cite{bai2025qwen3vltechnicalreport}. The results suggest that late-layer override is not tied to a specific architecture. Our contributions are as follows:

\begin{itemize}[leftmargin=*]
 \item We introduce Modal Dominance Ratio to trace layer-wise  modal preference and uncover late-layer textual override as a characteristic failure pattern. We show that transition direction,  not mere occurrence, predicts correctness.
 \item We propose CALRD, a training-free method that detects harmful overrides through complementary signals and applies adaptive correction, preserving beneficial processing.
 \item Experiments show CALRD achieves up to 9.4\% gains on conflict tasks while largely preserving standard performance. We also contribute Conflict-VQA, a diagnostic benchmark with explicit visual-textual annotations for mechanistic analysis.
\end{itemize}

\begin{figure}[t]
\centering
  \includegraphics[width=\linewidth]{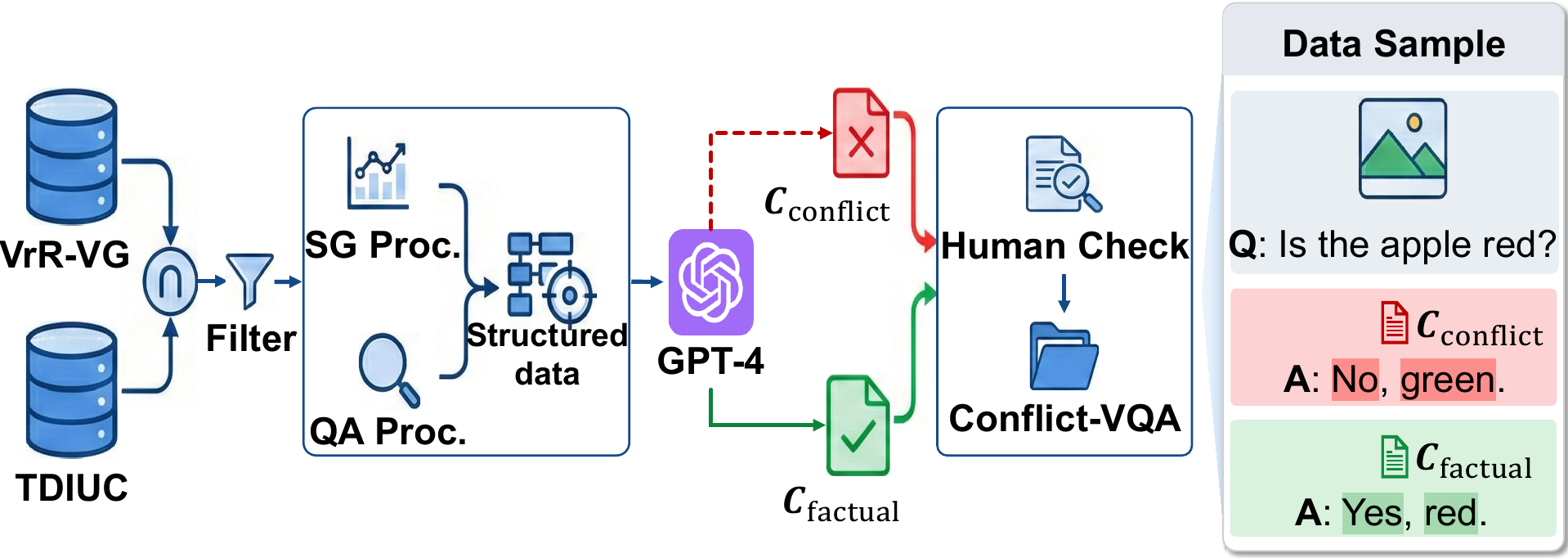}
\caption{Conflict-VQA construction pipeline. We process images from VrR-VG and questions from TDIUC, use GPT-4o to generate $C_{\text{factual}}$ and $C_{\text{conflict}}$, and verify all samples manually.}
  \label{fig:data_pipeline}
\end{figure}

\begin{table}[t]
\centering
\resizebox{\columnwidth}{!}{
\begin{tabular}{ccccc}
\toprule
InstructBLIP & LLaVA-1.5 & LLaVA-1.6 & Qwen2.5-VL & Qwen3-VL \\
\midrule
2,017 & 1,212 & 2,506 & 3,940 & 5,252 \\
\bottomrule
\end{tabular}
}
\caption{Competent subset size per model. Total samples: 5,969. A sample is competent if the model answers correctly in all five trials under non-conflict context.}
\label{tab:data_stats}
\end{table}

\section{Related Work}

\paragraph{Multimodal knowledge conflicts.} When information sources disagree, which should a model trust? Prior work has explored this question for text-only language models~\cite{Rich_Knowledge_Sources,xu-etal-2024-knowledge-conflicts,xie2023adaptive}. The problem becomes more pronounced in multimodal settings, where visual perception and textual claims can point to different answers. A number of recent benchmarks document this challenge. HallusionBench~\cite{Hallusionbench}, AutoHallusion~\cite{wu2024autohallusion}, and PhD~\cite{PhD_ChatGPT} report that MLLMs often follow textual context even when it contradicts what is visible in the image. Subsequent analyses suggest this behavior is systematic rather than accidental: models tend to rely on language priors or internal knowledge when facing conflicting evidence~\cite{Insight_Over_Sight,nguyen2025challenges,deng2025words}. More targeted benchmarks explicitly construct multimodal conflicts and measure which modality dominates, further confirming this tendency~\cite{jia2025benchmarking,zhang2025evaluating,zhu2024crossmodalityconflict}. While these studies clearly establish the phenomenon, they mostly describe it at the output level without explaining how conflicts are resolved inside the model. Our work addresses this gap by examining how visual and textual signals interact across layers, showing that many failures occur after visual information has already been encoded correctly.

\paragraph{Hallucination mitigation in MLLMs.} Various inference-time methods have been proposed to reduce hallucinations. Contrastive decoding is a common strategy: VCD suppresses tokens preferred under distorted images~\cite{liu2023mitigating}, ICD contrasts standard versus perturbed instructions~\cite{icd2024}, and OPERA penalizes attention over-trust patterns~\cite{huang2024opera}. Recent variants refine these ideas through explicit attention steering~\cite{wang2025ascd}, multi-stage contrast with selective visual inputs~\cite{park2025second}, and probabilistic detection that avoids a second forward pass~\cite{fieback2025ecd}. A separate line of work exploits model depth. DoLa compares predictions from shallow and deep layers to recover factual signals~\cite{dola2024}. DeCo, SHIFT, and related decoders rely on the observation that intermediate layers preserve stronger visual grounding~\cite{wang2024deco,wang2025shift,layercd_2025}. Training-based approaches reduce hallucinations through preference optimization or contrastive learning but require supervision~\cite{wang-etal-2024-mdpo,Yu_2024_CVPR,jiang2024hallucination}. Most methods correct uniformly, without distinguishing helpful from harmful late-layer processing. Our finding that transition \emph{direction} predicts correctness enables selective intervention only when needed. As we show in Section~\ref{sec:experiments}, uniform correction methods like DoLa can degrade performance when applied indiscriminately.

\paragraph{Layer-wise Analysis of Multimodal Models.} Recent studies examine visual information flow in MLLMs, finding that visual signals peak in intermediate layers and weaken in deeper ones~\cite{wang2024valid,knoweledge_evolve2025,yin2025lifting}. Others identify failure modes such as attention drift and modality imbalance that reduce visual grounding~\cite{jiang2025devils,chen2025multimodal}. These findings help explain why models capture visual details early but produce text-driven answers at the end. Our work extends this line of research. We show that the \emph{direction} of layer-wise transitions predicts final correctness, allowing us to intervene only when late-layer processing undermines visual evidence.

\begin{figure}[t]
\centering
  \includegraphics[width=\linewidth]{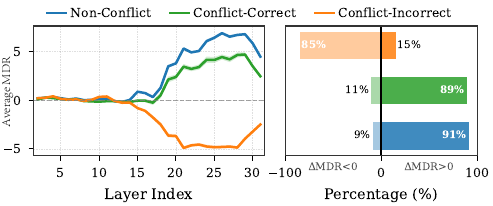}
  \caption{Layer-wise MDR analysis on LLaVA-1.6-7B. \textbf{Left:} Non-Conflict and Conflict-Correct samples maintain positive MDR, while Conflict-Incorrect samples reverse from positive to negative in late layers. \textbf{Right:} Text-ward shifts ($\Delta$MDR$<$0) dominate failures (85\%), vision-ward shifts ($\Delta$MDR$>$0) dominate correct predictions (89--91\%). Shading: $\pm$1 std.}
  \label{fig:mdr_trajectory}
\end{figure}

\section{Diagnosing Late-Layer Textual Override}
\label{sec:analysis}

To understand why models favor text over vision under conflict, we need to trace how their predictions evolve during the forward pass. Does textual bias arise because visual information is never properly encoded? Or is it encoded but subsequently discarded? This section presents a layer-wise analysis that reveals a surprising answer.

\subsection{Experimental Framework}
\label{sec:setup}

\paragraph{Task and data.} We study visual question answering where inputs consist of image $I$, context $C$, and question $Q$. Let $w_{\text{vis}}$ and $w_{\text{text}}$ denote answers supported by the image and context respectively. A \emph{knowledge conflict} occurs when $w_{\text{vis}} \neq w_{\text{text}}$. We categorize samples into three groups based on context type and model output. \textbf{Non-Conflict}: the context is consistent with the image and the model answers correctly. \textbf{Conflict-Correct}: context contradicts image but the model follows visual evidence. \textbf{Conflict-Incorrect}: the context contradicts the image and the
model follows textual claims.

Existing conflict benchmarks lack the $(w_{\text{vis}}, w_{\text{text}})$ annotations required for layer-wise analysis. To address this, we construct \textbf{Conflict-VQA}: 5,969 samples with paired consistent and conflicting contexts. We source images from VrR-VG~\cite{VrR-VG2019} and questions from TDIUC~\cite{TD-IUC2017}, covering six question types: object
presence, color, attribute, counting, spatial reasoning, and activity recognition. All contexts
are generated using GPT-4o and manually verified. Construction details are provided in Appendix A. For mechanistic analysis, we filter each model's competent subset, defined as samples answered correctly under consistent context across five trials. This filtering ensures that observed failures under conflict reflect conflict resolution process rather than basic visual perception errors.

\paragraph{Layer-wise projection.} Following early-exit methods~\cite{depth_adaptive2020,confident_adaptive2022}, we project hidden states at each layer $l$ to the output vocabulary via the language model head $W_{\text{head}}$: $P^{(l)}_t=\text{softmax}(W_{\text{head}}h^{(l)}_t)$, where $h^{(l)}_t$ is the hidden state at the answer token position and $P^{(l)}_t(w)$ is the scalar probability assigned to vocabulary token $w$. This lets us observe the model's ``preference'' at each depth.

\begin{figure}[t]
\centering
  \includegraphics[width=\linewidth]{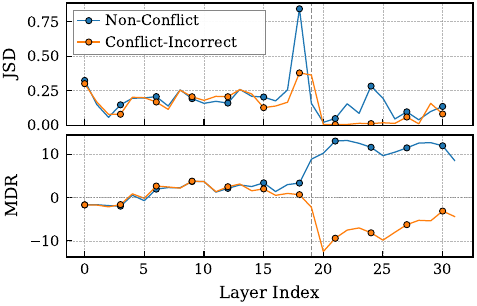}
  \caption{JSD and MDR trajectories on LLaVA-1.6-7B. JSD spikes align across categories, while MDR shifts toward vision in Non-Conflict and toward text in Conflict-Incorrect.}
  \label{fig:jsd_mdr}
\end{figure}

\subsection{Visual Predictions Emerge, Then Fade}
\label{sec:discovery1}

When models fail under conflict, is visual information absent from their representations, or present but ultimately ignored?

To answer this, we introduce \textbf{Modal Dominance Ratio (MDR)}, quantifying modal preference at each layer:
\begin{equation}
\text{MDR}^{(l)} = \log \frac{P^{(l)}_t(w_{\text{vis}})}{P^{(l)}_t(w_{\text{text}})}
\end{equation}
Positive MDR indicates preference for the visual answer; negative indicates textual preference.

Figure~\ref{fig:mdr_trajectory} shows MDR trajectories averaged across sample categories.
Non-Conflict and Conflict-Correct samples maintain positive MDR throughout all layers, reflecting consistent visual preference. Conflict-Incorrect samples tell a different story: \textbf{MDR is positive in early and middle layers, then reverses to negative in late layers.}

This pattern indicates that in failure cases, models assign higher probability to the visual answer in intermediate layers before shifting toward the textual answer. In other words, the visual information is present in intermediate representations but does not survive to the final output. We term this phenomenon \emph{late-layer textual override}. This observation suggests that the failure mechanism lies in late-layer processing rather than in early-layer encoding.

\subsection{Transition Direction Predicts Success}
\label{sec:discovery2}

The late-layer reversal in Conflict-Incorrect samples might suggest a straightforward fix: simply use intermediate-layer predictions. But prediction transitions also occur when models answer correctly. In these cases, late-layer processing appears to refine the answer rather than override it. To intervene selectively, we need a way to tell these two situations apart.

\begin{figure}[t]
\centering
    \includegraphics[width=\linewidth]{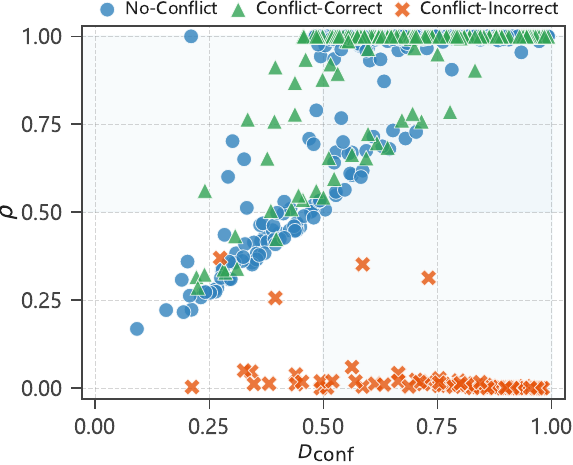}
    \caption{Detection signal distribution on LLaVA-1.6-7B. Conflict-Incorrect samples (red) exhibit high $D_{\mathrm{conf}}$ but low $\rho$. Non-Conflict (blue) and Conflict-Correct (green) maintain high $\rho$.}
    \label{fig:detection_signals}
\end{figure}

\paragraph{Detecting transitions.} Layer-wise predictions show high entropy in early layers and lower entropy in deeper layers as predictions converge.
We define the \emph{stability onset} $L_{\text{start}}$ as the layer with the largest entropy drop:
\begin{equation}
L_{\text{start}} = \arg\max_l [H_t^{(l-1)} - H_t^{(l)}],
\end{equation}
where $H_t^{(l)} = -\sum_w P^{(l)}_t(w) \log P^{(l)}_t(w)$.
Across our models, $L_{\text{start}}$ falls between 25\% and 40\% of model depth.

After stability onset, we detect abrupt distributional shifts using Jensen-Shannon divergence between adjacent layers: \begin{equation}
\text{JSD}^{(l)} = \frac{1}{2}\left[D_{\text{KL}}(P^{(l)}_t \| M) + D_{\text{KL}}(P^{(l+1)}_t \| M)\right],
\end{equation}
where $M = \frac{1}{2}(P^{(l)}_t + P^{(l+1)}_t)$. The transition layer is: $L_{\text{trans}} = \arg\max_{l \in [L_{\text{start}}, N-1]} \text{JSD}^{(l)}$.

\paragraph{Transition direction distinguishes success from failure.} Figure~\ref{fig:jsd_mdr} shows that both Non-Conflict and Conflict-Incorrect samples have JSD spikes at similar depths, so transitions alone do not distinguish the two groups. What does differ is the \emph{direction} of the shift. We quantify this via MDR change across the transition:
\begin{equation}
\Delta\text{MDR} = \overline{\text{MDR}}_{\text{after}} - \overline{\text{MDR}}_{\text{before}}
\end{equation}
where bars denote averages over pre/post-transition layers.

\paragraph{Finding.} In 85\% of Conflict-Incorrect cases, $\Delta\text{MDR}$ is negative, indicating a shift toward text. The pattern reverses for correct predictions: 89\% of Conflict-Correct and 91\% of Non-Conflict samples show positive $\Delta\text{MDR}$, shifting toward vision or holding steady. Transitions occur in all groups, but only the direction correlates with correctness.

\begin{figure}[t]
\centering
  \includegraphics[width=\linewidth,height=0.45\linewidth]{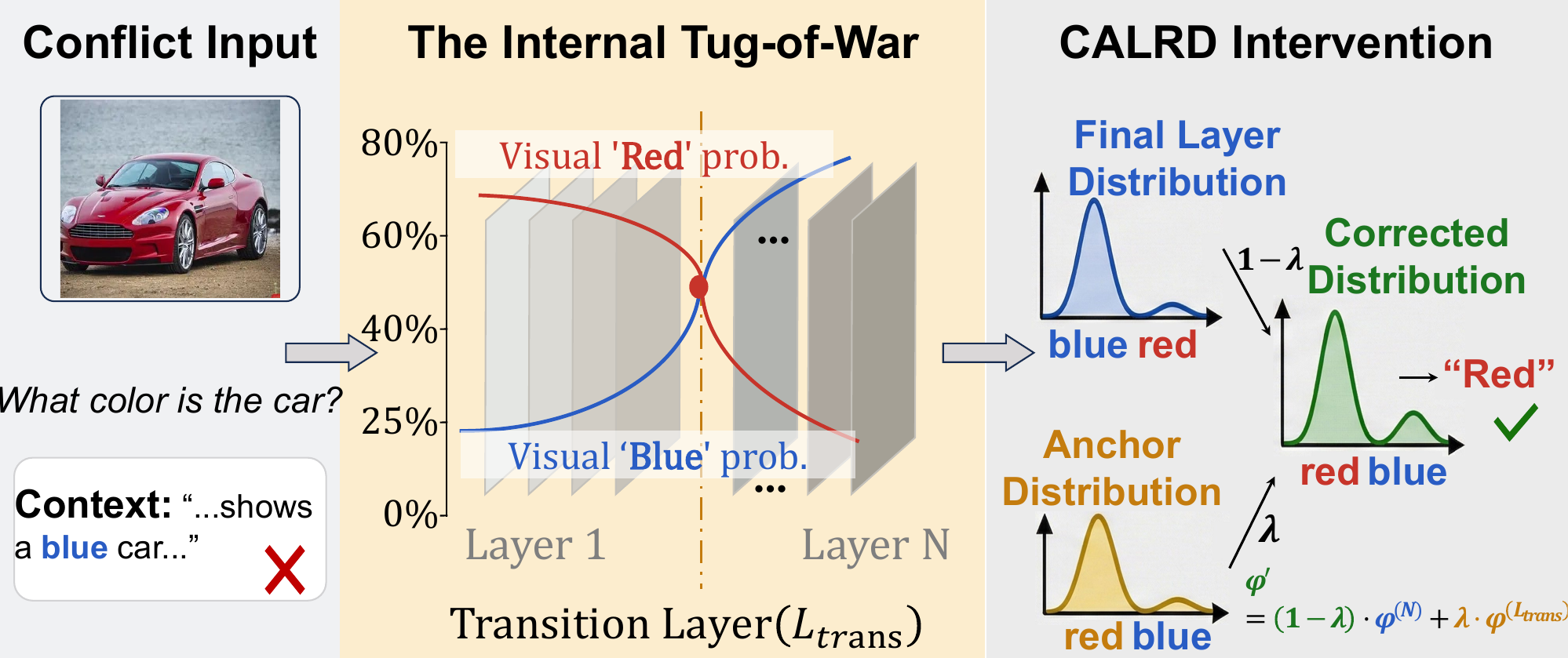}
  \caption{Overview of CALRD. Under visual-textual conflict, predictions shift from vision toward text across layers. CALRD detects this override and restores the transition-layer prediction.}
\end{figure}

\section{Conflict-Aware Layer Reference Decoding}
\label{sec:method}

The analysis in Section~\ref{sec:analysis} shows that harmful overrides are characterized by a late-layer shift from visual to textual preference, and that the direction of this shift distinguishes failures from successes. However, the MDR analysis relies on ground-truth $(w_{\text{vis}}, w_{\text{text}})$ labels, unavailable at inference. Can we detect harmful overrides without labels? In this section, we develop practical detection signals and a decoding strategy that recovers overridden predictions without requiring such labels.

Specifically, we propose \emph{Conflict-Aware Layer Reference Decoding} (CALRD), a training-free method. CALRD first identifies when a confident transition-layer prediction is being suppressed, then adjusts the output distribution to restore it. Since prediction transitions also occur in correct outputs, the intervention must be selective: we modulate its strength based on how strongly the override signature is present.

\subsection{Detecting Harmful Override}

The MDR analysis shows that Conflict-Incorrect samples hold confident predictions at $L_{\text{trans}}$ that do not survive to the final layer (section~\ref{sec:analysis}). This observation motivates two simple signals. Let $w^* = \arg\max_w P^{(L_{\text{trans}})}_t(w)$ be the top prediction at transition. Our detection signals are:

\paragraph{Anchor confidence.} A higher transition-layer probability for $w^*$ indicates a stronger intermediate preference:
\begin{equation}
D_{\text{conf}} = P^{(L_{\text{trans}})}_t(w^*)
\end{equation}

\paragraph{Prediction retention.} To quantify survival, we use the final-to-transition probability ratio; $\rho \approx 1$ indicates preservation, while $\rho \approx 0$ indicates override:
\begin{equation}
\rho = \min\left(1, \frac{P^{(N)}_t(w^*)}{P^{(L_{\text{trans}})}_t(w^*)}\right)
\end{equation}

Figure~\ref{fig:detection_signals} shows how these signals distribute across sample types. Conflict-Incorrect samples cluster in the high-$D_{\text{conf}}$, low-$\rho$ region: confident predictions that get overridden. Non-Conflict and Conflict-Correct samples maintain high $\rho$, meaning their predictions persist. This separation allows us to detect harmful overrides at test time.

\subsection{Adaptive Intervention}

Not all prediction transitions are harmful. As shown in Section~\ref{sec:discovery2}, transitions toward vision correlate with correct outputs, while transitions toward text correlate with failures. To avoid disrupting beneficial late-layer processing, intervention strength must adapt to override severity. We define:

\begin{equation}
\lambda = D_{\text{conf}} \cdot (1 - \rho)
\label{eq:lambda}
\end{equation}

The multiplicative form ensures that $\lambda$ is substantial only when \emph{both} conditions hold: a confident transition-layer prediction exists ($D_{\text{conf}}$ high) \emph{and} it is being suppressed ($\rho$ low). When $D_{\text{conf}}$ is low, there is no preference to recover, so $\lambda$ stays small regardless of retention. When $\rho$ is high, the prediction survives to the output, so intervention is unnecessary. Only when a confident prediction gets discarded does $\lambda$ become large enough to affect decoding. This design is motivated by the asymmetry in Section~\ref{sec:discovery2}: harmful cases are likely to exhibit confident intermediate predictions that are later suppressed, whereas correct cases tend to preserve them.

\subsection{Transition-Layer Guided Decoding}

Given $\lambda$, we adjust the output logits by combining final-layer and transition-layer predictions. Let $\phi^{(l)}_t = W_{\text{head}} \cdot h^{(l)}_t$ denote the unnormalized logits at layer $l$. The adjusted logits are:
\begin{equation}
\phi'_t = (1 - \lambda) \cdot \phi^{(N)}_t + \lambda \cdot \phi^{(L_{\text{trans}})}_t
\end{equation}
with output distribution $P'_t = \text{softmax}(\phi'_t)$. When $\lambda = 0$, this reduces to standard decoding. As $\lambda$ increases, the transition-layer prediction receives more weight, recovering the vision-grounded answer that would otherwise be lost. For samples without override signatures, $\lambda$ remains near zero and the output is largely unchanged.

CALRD works at each token position independently. For multi-token generation, we recompute $L_{\text{trans}}$, $D_{\text{conf}}$, and $\rho$ at each step, allowing the intervention strength to vary across the sequence. The logit-level operation makes CALRD compatible with greedy, beam search, and nucleus sampling without modification. Note that CALRD does not introduce external knowledge or require additional training. The correction comes entirely from the model's own intermediate representations. As our analysis shows, the vision-grounded answer is often already present at the transition layer; CALRD simply helps it survive to the output.

\begin{table*}[t!]
\centering
\small
\resizebox{\linewidth}{!}{
    \begin{tabular}{ll|ccccc|ccccc}
    \toprule
    & & \multicolumn{5}{c|}{\textbf{Conflict-VQA (Acc~$\uparrow$)}} & \multicolumn{5}{c}{\textbf{PhD-icc (Acc~$\uparrow$)}} \\
    \textbf{Decoding} & \textbf{Method} & InstructBLIP & LLaVA-1.5 & LLaVA-1.6 & Qwen2.5-VL & Qwen3-VL & InstructBLIP & LLaVA-1.5 & LLaVA-1.6 & Qwen2.5-VL & Qwen3-VL \\
    \midrule
    \multirow{4}{*}{Greedy}
     & Vanilla & \num{39.61} & \num{36.20} & \num{42.35} & \num{65.61} & \num{75.58} & \num{41.08} & \num{27.72} & \num{28.23} & \num{51.30} & \num{60.32} \\
     & DoLa    & \num{32.72} & \num{40.10} & \num{44.59} & \num{66.22} & \num{76.06} & \num{32.65} & \num{29.78} & \num{28.73} & \num{51.08} & \num{58.00} \\
     & DeCo    & \num{41.31} & \num{38.03} & \num{43.38} & \num{63.79} & \num{75.33} & \num{42.12} & \num{31.00} & \num{33.65} & \num{51.65} & \num{62.25} \\
     \rowcolor{bestrow}
     &\textbf{ CALRD (Ours) }& \resg{44.12}{4.5} & \resg{41.90}{5.7} & \resg{49.51}{7.2} & \resg{70.15}{4.5} & \resg{77.42}{1.8} & \resg{49.62}{8.5} & \resg{36.35}{8.6} & \resg{36.52}{8.3} & \resg{56.20}{4.9} & \resg{63.68}{3.4} \\
    \midrule
    \multirow{4}{*}{Beam}
     & Vanilla & \num{39.98} & \num{38.88} & \num{43.99} & \num{65.49} & \num{75.94} & \num{40.73} & \num{28.25} & \num{28.80} & \num{51.11} & \num{62.53} \\
     & OPERA   & \num{40.12} & \num{39.00} & \num{43.74} & \num{68.49} & \num{76.43} & \num{39.78} & \num{28.93} & \num{29.95} & \num{53.70} & \num{62.62} \\
     & DeCo    & \num{40.83} & \num{38.15} & \num{43.01} & \num{63.55} & \num{75.46} & \num{42.15} & \num{31.07} & \num{34.63} & \num{51.13} & \num{61.25} \\
     \rowcolor{bestrow}
     & \textbf{CALRD (Ours) }& \resg{44.49}{4.5} & \resg{44.47}{5.6} & \resg{50.79}{6.8} & \resg{70.25}{4.8} & \resg{79.34}{3.4} & \resg{49.75}{9.0} & \resg{36.58}{8.3} & \resg{37.40}{8.6} & \resg{56.70}{5.6} & \resg{64.60}{2.1} \\
    \midrule
    \multirow{4}{*}{Nucleus}
     & Vanilla & \num{23.94} & \num{38.76} & \num{41.92} & \num{61.48} & \num{73.39} & \num{39.23} & \num{29.33} & \num{30.08} & \num{51.05} & \num{62.23} \\
     & VCD     & \num{23.05} & \num{42.40} & \num{46.35} & \num{64.04} & \num{75.50} & \num{40.01} & \num{31.87} & \num{31.70} & \num{53.83} & \num{63.40} \\
     & DeCo    & \num{24.91} & \num{38.45} & \num{45.44} & \num{63.55} & \num{74.79} & \num{41.83} & \num{31.25} & \num{32.90} & \num{51.62} & \num{61.50} \\
     \rowcolor{bestrow}
     &\textbf{ CALRD (Ours)} & \resg{28.07}{4.1} & \resg{43.40}{4.6} & \resg{48.72}{6.8} & \resg{68.91}{7.4} & \resg{76.55}{3.2} & \resg{48.62}{9.4} & \resg{34.95}{5.6} & \resg{34.12}{4.0} & \resg{55.83}{4.8} & \resg{64.52}{2.3} \\
    \bottomrule
    \end{tabular}
}
\caption{Performance comparison on knowledge conflict benchmarks, where textual context contradicts visual evidence. We evaluate five MLLMs with three decoding strategies. Best results are in \textbf{bold}. {\footnotesize\textcolor{red}{↑}}: improvement over Vanilla.}
\label{tab:conflict}
\end{table*}

\section{Experiments}
\label{sec:experiments}

We evaluate CALRD on two questions: (1) Does it improve conflict resolution? (2) Does it hurt non-conflict scenarios? Experiments cover five MLLMs with different architectures and capability levels. For each model, we report results on conflict benchmarks where context contradicts the image, as well as standard hallucination benchmarks.

\subsection{Experimental Setup}

\paragraph{Models.} We evaluate five MLLMs: InstructBLIP-7B~\cite{instructblip2023}, LLaVA-1.5-7B, LLaVA-1.6-7B~\cite{llava-1.5}, Qwen2.5-VL-7B~\cite{bai2025qwen2}, and Qwen3-VL-8B~\cite{bai2025qwen3vltechnicalreport}. These models span different architectures and capability levels.

\paragraph{Benchmarks.} We evaluate on two complementary benchmark groups. (1) For conflict resolution, we use \textbf{Conflict-VQA} (Section~\ref{sec:setup}), which contains 5,969 samples requiring open-ended answers, and \textbf{PhD-icc}~\cite{PhD_ChatGPT}, the incorrect-context subset of PhD containing 16,844 yes/no question pairs. Both benchmarks include contexts that contradict visual evidence, and we report accuracy. (2) For standard hallucination without explicit conflicts, we use \textbf{POPE}~\cite{pope2023} and \textbf{CHAIR}~\cite{Object_Hallucination}. POPE probes object hallucination by asking ``Is there a \texttt{<object>} in the image?'' across random, popular, and adversarial splits. It covers 500 MSCOCO images with six questions per image for each split, and we report F1 score. CHAIR measures caption hallucination against ground-truth object annotations, reporting instance-level $\text{C}_I$ and sentence-level $\text{C}_S$ (lower is better). Following~\cite{huang2024opera}, we evaluate on 500 MSCOCO images with the same captioning prompt ``Please describe this image in detail.'' for consistency.

\paragraph{Baselines.} We compare against four representative methods. DoLa~\cite{dola2024} contrasts logits from later layers against earlier layers to surface factual knowledge. VCD~\cite{vcd2024} contrasts outputs from original and distorted visual inputs to reduce over-reliance on language priors. OPERA~\cite{huang2024opera} penalizes over-trust patterns in self-attention and uses rollback to re-select tokens. DeCo~\cite{wang2024deco} adaptively selects preceding layers where visual information is stronger and integrates their predictions into the final output. We test CALRD with greedy, beam search, and nucleus sampling. Since not all baselines are designed for every decoding strategy, we evaluate each method under the decoding regimes supported by its official implementation and compare CALRD against the same vanilla decoder in each regime. All experiments are run on NVIDIA H100 GPUs.

\subsection{Main Results}

\paragraph{Conflict resolution.} Table~\ref{tab:conflict} shows results on conflict benchmarks. CALRD outperforms all baselines across models and decoders. Gains are largest on PhD-icc: LLaVA-1.5 improves from 27.72\% to 36.35\% (+8.63\%), LLaVA-1.6 from 28.23\% to 36.52\% (+8.29\%), and InstructBLIP from 41.08\% to 49.62\% (+8.54\%). On Conflict-VQA, improvements reach up to 7.4\%.

\paragraph{Larger gains on PhD-icc.} The two benchmarks differ in question format: PhD-icc uses binary yes/no questions, while Conflict-VQA requires open-ended answers. We suspect the binary format creates conditions where textual bias is more pronounced. When context strongly suggests ``yes'' or ``no,'' the model may be more easily swayed away from visual evidence. In contrast, open-ended questions require generating specific content, making it harder for context alone to override perception.

\paragraph{Baseline accuracy and improvement.} An interesting pattern emerges when we rank models by their vanilla accuracy on PhD-icc: LLaVA-1.5 (27.72\%) $<$ LLaVA-1.6 (28.23\%) $<$ InstructBLIP (41.08\%) $<$ Qwen2.5-VL (51.30\%) $<$ Qwen3-VL (60.32\%). The gains from CALRD roughly follow the inverse order, with weaker models seeing larger improvements. This pattern is expected: stronger models already resolve more conflicts correctly in their late layers, leaving fewer overridden predictions for CALRD to recover. The adaptive nature of our intervention naturally accommodates this---when predictions survive to the output ($\rho$ remains high), $\lambda$ stays small and CALRD leaves the output largely unchanged. For Qwen3-VL, the +3.36\% absolute gain corresponds to an 8.5\% relative error reduction, indicating that even well-calibrated models contain recoverable failures.

\paragraph{Comparison with other layer-based methods.} DeCo and DoLa also leverage information from earlier layers, but their performance is less consistent. DoLa improves some configurations but hurts others: on InstructBLIP, Conflict-VQA accuracy drops from 39.61\% to 32.72\%. DeCo shows more stable behavior but still underperforms CALRD across the board. One possible explanation is that these methods apply corrections more uniformly, without distinguishing cases where late-layer processing is beneficial. CALRD's detection signals allow it to intervene selectively, reducing the risk of disrupting predictions that were already on track.

\paragraph{Non-conflict scenarios.} Textual override is not identical to all hallucinations, so we also test CALRD on standard object-centric hallucination benchmarks without explicit conflicts. A method that improves conflict resolution but degrades standard performance would have limited practical value. Tables~\ref{tab:pope_results} and~\ref{tab:chair_results} show that CALRD largely avoids this trade-off.

\paragraph{POPE results (Table~\ref{tab:pope_results}).} CALRD maintains or improves F1 across all configurations. Under greedy decoding, InstructBLIP improves from 80.0 to 85.4, and Qwen3-VL from 88.5 to 91.6. POPE probes object hallucination without explicit visual-textual conflicts, so these results suggest that late-layer override may occur more broadly than just in conflict settings. The key is that CALRD does not intervene indiscriminately: when $\rho$ is high, indicating the prediction is stable, $\lambda$ stays near zero and the output remains unchanged.

\paragraph{CHAIR results (Table~\ref{tab:chair_results}).} Captioning differs from VQA in that it requires extended sequences rather than short answers. Hallucinations accumulate over multiple tokens, making it a useful stress test. CALRD reduces both $\text{C}_I$ and $\text{C}_S$ in most configurations. InstructBLIP shows the largest improvement: $\text{C}_I$ drops from 23.7 to 14.6 (38\% relative reduction) and $\text{C}_S$ from 58.8 to 40.2. These results indicate that recomputing detection signals at each token position allows CALRD to adapt throughout the sequence, rather than a fixed correction.

\begin{table}[t!]
\centering
\resizebox{\linewidth}{!}{
    \begin{tabular}{ll|ccccc}
    \toprule
    & & \multicolumn{5}{c}{\textbf{POPE (F1~$\uparrow$)}} \\
    \cmidrule(l){3-7}
    \textbf{Dec.} & \textbf{Method} & \textbf{InstructBLIP} & \textbf{LLaVA-1.5} & \textbf{LLaVA-1.6} & \textbf{Qwen2.5-VL} & \textbf{Qwen3-VL} \\
    \midrule
    \multirow{4}{*}{Greedy}
     & Vanilla & \num{80.0} & \num{82.2} & \num{85.4} & \num{80.8} & \num{88.5} \\
     & DoLa    & \num{83.4} & \num{83.2} & \num{87.1} & \num{82.1} & \num{88.6} \\
     & DeCo    & \num{84.9} & \num{83.8} & \num{87.6} & \num{81.4} & \num{90.6} \\
      \rowcolor{bestrow}
     & \textbf{CALRD (Ours)} & \resup{85.4}{5.4} & \resup{84.5}{2.3} & \resup{88.2}{2.8} & \resup{82.3}{1.5} & \resup{91.6}{3.1} \\
    \midrule
    \multirow{4}{*}{Beam}
     & Vanilla & \num{84.4} & \num{84.9} & \num{87.1} & \num{80.7} & \num{88.6} \\
     & OPERA   & \num{84.8} & \num{85.4} & \num{87.5} & \num{80.8} & \num{89.3} \\
     & DeCo    & \num{84.9} & \num{86.7} & \num{87.9} & \num{81.4} & \num{91.2} \\
      \rowcolor{bestrow}
     & \textbf{CALRD (Ours)} & \resup{84.7}{0.3} & \resup{86.9}{2.0} & \resup{88.3}{1.2} & \resup{82.6}{1.9} & \resup{91.9}{3.3} \\
    \midrule
    \multirow{4}{*}{Nucleus}
     & Vanilla & \num{79.8} & \num{83.1} & \num{85.5} & \num{80.8} & \num{88.7} \\
     & VCD     & \num{79.9} & \num{83.1} & \num{85.6} & \num{80.9} & \num{88.9} \\
     & DeCo    & \num{81.8} & \num{84.8} & \num{87.3} & \num{81.3} & \num{91.2} \\
      \rowcolor{bestrow}
     & \textbf{CALRD (Ours)} & \resup{83.4}{3.6} & \resup{85.9}{2.8} & \resup{87.5}{2.0} & \resup{81.9}{1.1} & \resup{91.4}{2.7} \\
    \bottomrule
    \end{tabular}
}
\caption{Results on POPE object hallucination benchmark. CALRD results are highlighted. {\footnotesize\textcolor{red}{↑}}: improvement over Vanilla.}
\label{tab:pope_results}
\end{table}

\begin{table*}[t]
\centering
\small
\resizebox{\linewidth}{!}{
    \begin{tabular}{ll|cc|cc|cc|cc|cc}
    \toprule
    & & \multicolumn{2}{c|}{\textbf{InstructBLIP}} & \multicolumn{2}{c|}{\textbf{LLaVA-1.5}} & \multicolumn{2}{c|}{\textbf{LLaVA-1.6}} & \multicolumn{2}{c|}{\textbf{Qwen2.5-VL}} & \multicolumn{2}{c}{\textbf{Qwen3-VL}} \\

    \textbf{Decoding} & \textbf{Method} & $\text{C}_S \downarrow$ & $\text{C}_I \downarrow$ & $\text{C}_S \downarrow$ & $\text{C}_I \downarrow$ & $\text{C}_S \downarrow$ & $\text{C}_I \downarrow$ & $\text{C}_S \downarrow$ & $\text{C}_I \downarrow$ & $\text{C}_S \downarrow$ & $\text{C}_I \downarrow$ \\

    \midrule

    \multirow{4}{*}{Greedy}

     & Vanilla & \numb{58.8} & \num{23.7} & \num{45.0} & \num{14.7} & \num{37.6} & \num{12.7} & \num{40.8} & \num{9.7} & \num{52.4} & \num{10.1} \\

     & DoLa    & \numb{48.4} & \num{15.9} & \num{47.8} & \num{13.8} & \num{35.3} & \num{8.7} & \num{36.5} & \num{9.2} & \num{55.6} & \num{9.8} \\

     & DeCo    & \numb{41.2} & \num{14.4} & \num{37.8} & \num{11.1} & \num{35.8} & \num{10.4} & \num{37.6} & \textbf{\num{9.1}} & \num{51.2} & \num{9.6} \\
 \rowcolor{bestrow}
     & \textbf{CALRD (Ours)} & \resdown{40.2}{18.6} & \resdown{14.6}{9.1} & \resdown{35.7}{9.3} & \resdown{10.3}{4.4} & \resdown{33.6}{4.0} & \resdown{9.3}{3.4} & \resdown{35.5}{5.3} & \resdown{9.5}{0.2} & \resdown{50.7}{1.7} & \resdown{9.3}{0.8} \\

    \midrule

    \multirow{4}{*}{Beam}

     & Vanilla & \numb{55.6} & \num{15.8} & \numb{48.8} & \num{13.9} & \num{36.0} & \num{12.1} & \num{38.9} & \num{9.4} & \numb{53.6} & \num{9.7} \\

     & OPERA   & \numb{46.4} & \num{14.2} & \numb{44.6} & \num{12.8} & \num{35.3} & \num{12.9} & \num{39.1} & \num{10.2} & \numb{54.7} & \num{11.3} \\

     & DeCo    & \numb{43.8} & \num{12.7} & \numb{32.0} & \num{9.7} & \num{34.4} & \num{9.0} & \num{38.2} & \num{8.5} & \numb{52.0} & \num{9.3} \\
 \rowcolor{bestrow}
     & \textbf{CALRD (Ours)} & \resdown{38.4}{17.2} & \resdown{11.4}{4.4} & \resdown{34.7}{14.1} & \resdown{10.9}{3.0} & \resdown{32.8}{3.2} & \resdown{8.7}{3.4} & \resdown{33.4}{5.5} & \resdown{8.2}{1.2} & \resdown{40.6}{13.0} & \resdown{7.9}{1.8} \\

    \midrule

    \multirow{4}{*}{Nucleus}

     & Vanilla & \numb{54.6} & \numb{24.8} & \numb{48.8} & \num{14.2} & \num{36.8} & \num{10.5} & \num{40.4} & \num{10.7} & \num{54.2} & \num{10.2} \\

     & VCD     & \numb{58.0} & \numb{17.0} & \numb{54.0} & \num{16.0} & \num{41.2} & \num{10.2} & \num{42.9} & \num{12.3} & \num{53.4} & \num{11.0} \\

     & DeCo    & \numb{43.6} & \numb{12.9} & \numb{42.8} & \num{13.2} & \num{36.1} & \num{10.1} & \num{40.4} & \num{9.5} & \num{54.8} & \num{9.9} \\
 \rowcolor{bestrow}
     & \textbf{CALRD (Ours)} & \resdown{42.7}{11.9} & \resdown{12.1}{12.7} & \resdown{37.6}{11.2} & \resdown{12.8}{1.4} & \resdown{34.8}{2.0} & \resdown{11.5}{-1.0} & \resdown{38.6}{1.8} & \resdown{9.1}{1.6} & \resdown{53.6}{0.6} & \resdown{9.5}{0.7} \\

    \bottomrule

    \end{tabular}

}

\caption{Results on CHAIR caption hallucination benchmark. Lower is better. {\footnotesize\textcolor{blue}{↓}}: reduction over Vanilla.}
\label{tab:chair_results}
\end{table*}

\begin{table}[t]
\centering
\begin{tabular}{l|cc}
\toprule
Configuration & C-VQA $\uparrow$ & POPE$\uparrow$  \\
\midrule
Vanilla (Greedy) & 42.3 & 85.4  \\
\midrule
CALRD (Full)     & \textbf{49.5} & \textbf{88.2}  \\
\quad w/o $D_{\text{conf}}$ & 47.3 & 87.5  \\
\quad w/o $\rho$  & 45.1 & 86.6  \\
\quad w/ Fixed Layer (75\%) & 45.9 & 87.1  \\
\bottomrule
\end{tabular}
\caption{Ablation study on LLaVA-1.6 with greedy decoding. We evaluate the contribution of each component.}
\label{tab:ablation}
\end{table}

\begin{figure}[t]
  \centering
  \includegraphics[width=\linewidth]{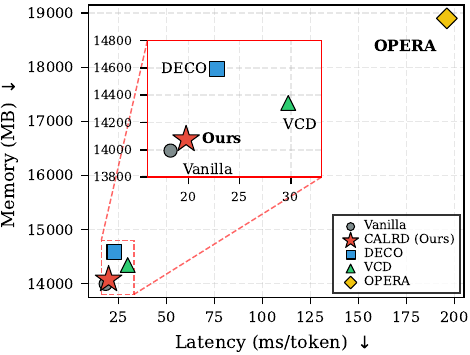}
  \caption{\textbf{Inference Efficiency on LLaVA-1.5-7B.} Latency vs. GPU memory trade-off measured on a single H100 GPU. Our CALRD achieves a superior balance, staying closest to the Vanilla baseline compared to DECO, VCD, and OPERA.}
  \label{fig:efficiency}
\end{figure}

\subsection{Analysis}

\paragraph{Ablation study.} Table~\ref{tab:ablation} examines the contribution of each component using LLaVA-1.6 with greedy decoding. Removing $D_{\text{conf}}$ drops Conflict-VQA accuracy from 49.5\% to 47.3\%, while removing $\rho$ causes a larger drop to 45.1\%. This suggests that retention is the more informative signal for detecting harmful override, which makes sense: a prediction can be confident at the transition layer for various reasons, but a sharp drop in retention specifically indicates that something changed in late-layer processing.

Using a fixed layer at 75\% depth instead of dynamic $L_{\text{trans}}$ detection reduces accuracy to 45.9\%. The value corresponds to the median transition layer observed in our analysis (Section~\ref{sec:discovery2}), so this comparison uses the most representative fixed position rather than an arbitrary choice. The accuracy drop indicates that transition points vary substantially across samples, and a single fixed layer cannot capture this variation. We also observe that POPE follows a similar pattern, with both signals contributing to the overall improvement.

These results support the multiplicative design $\lambda = D_{\text{conf}} \cdot (1 - \rho)$: both signals provide useful information, but neither is sufficient alone. The product ensures that substantial intervention occurs only when a confident prediction exists and is being suppressed by late-layer processing.

\paragraph{Efficiency.} Figure~\ref{fig:efficiency} compares latency and memory usage across methods using LLaVA-1.5 on a single H100 GPU. CALRD adds modest overhead: 19.8 ms/token compared to 18.3 ms/token for vanilla decoding, with only ~80 MB additional memory. VCD nearly doubles latency due to its dual-stream requirement, and OPERA is approximately \textbf{10$\times$} slower because of iterative rollback. DeCo falls in between. These results suggest CALRD is practical for deployment scenarios where efficiency matters.

\section{Conclusion}

We studied why multimodal models favor text over vision when the two conflict, and found that the problem is not poor visual encoding. In many failure cases, models assign higher probability to the correct visual answer in intermediate layers, only to override it with the textual answer in deeper layers. We call this late-layer textual override. Importantly, the direction of prediction change distinguishes failures from successes: shifts toward text correlate with incorrect outputs, while shifts toward vision correlate with correct ones. This observation motivates CALRD, a training-free decoding method that restores intermediate predictions when override signatures are detected. Experiments on five MLLMs show up to 9.4\% improvements on conflict benchmarks while largely preserving performance on standard tasks. Our results suggest that improving reliability under conflict may require not better visual encoding, but better preservation of visual information through the forward pass.

\paragraph{Limitations and future work.} Our work opens several avenues for future research. A natural next step is to examine how late-layer dynamics interact with model scale and pretraining data composition—questions that require controlled experiments with access to training corpora, beyond the scope of our mechanistic study. Additionally, exploring whether similar override patterns emerge across other modality pairs (\textit{e.g., }audio-text) may reveal general principles of multimodal integration. We view our mechanistic analysis as a foundation for such investigations.

\section*{Acknowledgments}

This work was supported by the National Natural Science Foundation of China under Grant 62306331, 62407037, and CAAI Youth Talent Lifting Project under Grant CAAI2023-2025QNRC001.

\nocite{fan2025visipruner}
\bibliographystyle{named}
\bibliography{ijcai26}

\clearpage
\def\SUPPLEMENTARYBODY{}
\appendix

\section*{Appendix Overview}

Due to space constraints in the main text, we provide supplementary materials in this appendix. Appendix A details the construction of Conflict-VQA, the diagnostic benchmark introduced in Section 3.1 for evaluating visual-textual conflict resolution. Appendix B provides the complete pseudocode for CALRD, complementing the method description in Section 4. Appendix C offers additional analysis of the late-layer textual override phenomenon, including a representative case study and discussion of potential causes. The remaining sections present additional experimental results (Appendix D), formal definitions of evaluation metrics (Appendix E), implementation details for reproducibility (Appendix F), and qualitative examples illustrating CALRD's behavior (Appendix G).

\section{Conflict-VQA: Benchmark Construction}
\label{sec:appendix_construction}

This section details the construction of Conflict-VQA, the diagnostic benchmark introduced in Section 3.1. Conflict-VQA is designed to evaluate how MLLMs resolve conflicts between visual evidence and textual context, with explicit annotations of both vision-grounded ($w_{\mathrm{vis}}$) and text-suggested ($w_{\mathrm{text}}$) answers.
\subsection{Data Sources and Preprocessing}

We construct Conflict-VQA by combining images from VrR-VG~\cite{VrR-VG2019} with questions from TDIUC~\cite{TD-IUC2017}, matching samples by \texttt{image\_id}. We focus on six question types where visual-textual conflicts can be unambiguously defined: object presence, color, attribute, counting, positional reasoning, and activity recognition.

VrR-VG provides scene graphs that we use to ground context generation. However, these annotations require preprocessing before use. Object identifiers are often inconsistent—the same object may appear under different names or be split into multiple entries. We normalize these by mapping synonyms to canonical names and merging duplicates while preserving count and attribute information. Relations are converted to readable \{\emph{subject}, \emph{predicate}, \emph{object}\} triples. This preprocessing yields a structured scene representation that grounds the subsequent context generation step.

\subsection{Answer Pair and Context Generation}

Each sample requires two answers: $w_{\mathrm{vis}}$ (the vision-grounded answer) and $w_{\mathrm{text}}$ (the text-suggested conflicting answer). The original TDIUC answer serves as $w_{\mathrm{vis}}$. We construct $w_{\mathrm{text}}$ based on question type: flipping yes/no for presence questions, substituting with a plausible alternative for color and attribute questions, perturbing counts by $\pm$1 or $\pm$2 for counting questions, flipping spatial relations for positional questions, and substituting with similar actions for activity questions. We discard cases where the constructed $w_{\mathrm{text}}$ would be implausible given the scene.

For each sample, we prompt GPT-4o to generate two contexts (see Figure~\ref{fig:prompt} for the complete prompt template). The factual context $C_{\mathrm{factual}}$ accurately describes the scene and supports $w_{\mathrm{vis}}$, while the conflict context $C_{\mathrm{conflict}}$ is nearly identical but contains exactly one statement supporting $w_{\mathrm{text}}$ instead. Both contexts are 3--5 sentences of natural prose, reference 2--3 real objects from the scene, use confident declarative language without hedging, and introduce no fabricated objects.

\subsection{Quality Control and Evaluation Protocol}

Two researchers independently verified all samples, checking that $C_{\mathrm{factual}}$ accurately describes the image, that $C_{\mathrm{conflict}}$ differs only in the intended contradictory statement, and that both contexts are fluent and natural. Disagreements were resolved through discussion, resulting in approximately 8\% of initial candidates being filtered out.

Table~\ref{tab:accuracy_relationship} reports three metrics for each model. Non-conflict accuracy measures baseline performance under $C_{\mathrm{factual}}$. Conflict accuracy measures performance under $C_{\mathrm{conflict}}$; the gap between the two reflects how much the misleading context hurts. The competent ratio is stricter: it requires a model to answer correctly in all five trials under $C_{\mathrm{factual}}$. We use the competent subset for mechanistic analysis in Section 3, ensuring that studied failures reflect conflict resolution rather than poor visual perception. The main-paper evaluation uses the full dataset.

\begin{figure*}[t]
  \centering
  \includegraphics[width=\linewidth]{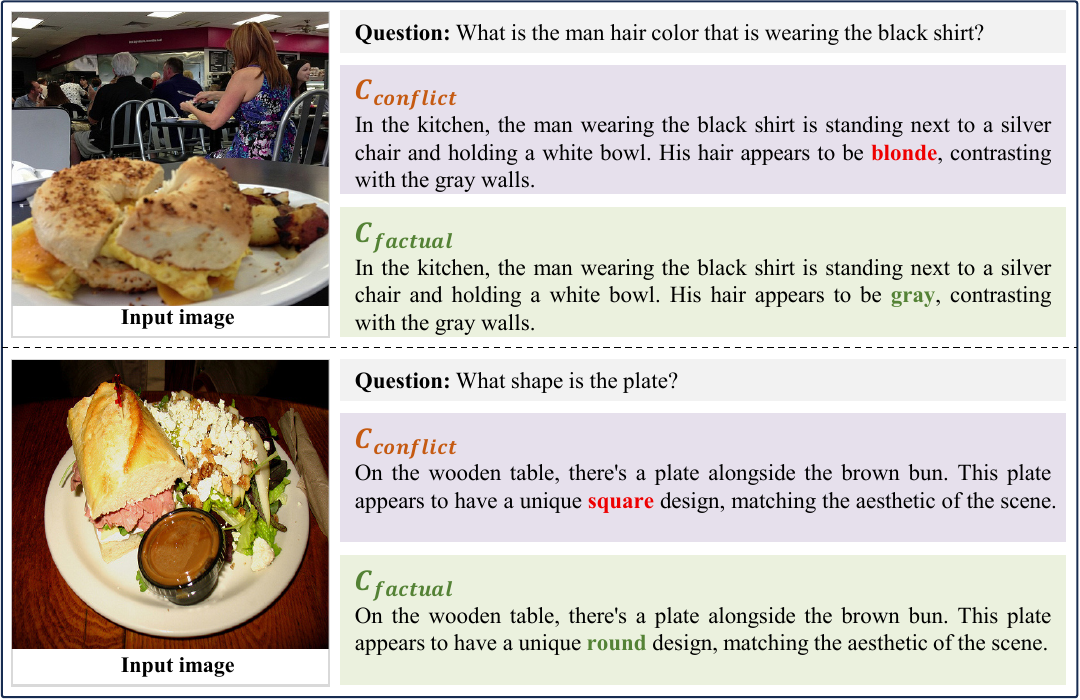}
  \caption{Examples from Conflict-VQA. $C_{\mathrm{conflict}}$ contains a false 
claim (red) that contradicts the image, while $C_{\mathrm{factual}}$ states the 
correct answer (green). Other parts of the two contexts remain identical.}
  \label{fig:data_example}
\end{figure*}

\begin{table}[t]
\centering

\begin{tabular}{lcccc}
\toprule
Model & Non-conf. & Conf. & \multicolumn{2}{c}{Competent} \\
\cmidrule(lr){4-5}
      & Acc. & Acc. & Ratio & $n$ \\
\midrule
InstructBLIP & 50.7 & 39.6 & 33.8 & 2,017 \\
LLaVA-1.5    & 38.7 & 36.2 & 20.3 & 1,212 \\
LLaVA-1.6    & 54.9 & 42.4 & 42.0 & 2,506 \\
Qwen2.5-VL   & 81.2 & 65.6 & 66.0 & 3,940 \\
Qwen3-VL     & 91.3 & 75.6 & 88.0 & 5,252 \\
\bottomrule
\end{tabular}
\caption{Accuracy under non-conflict and conflict conditions. Competent subset: correct in all five trials under $C_{\mathrm{factual}}$ ($N$=5,969).}
\label{tab:accuracy_relationship}
\end{table}

\section{CALRD Algorithm}
\label{sec:appendix_algorithm}

This section provides the complete pseudocode for Conflict-Aware Layer Reference Decoding (CALRD), the training-free decoding method proposed in Section 3.4. Algorithm~\ref{alg:calrd} details the three-stage procedure executed at each step.

\paragraph{Stage 1: Transition Layer Localization.} CALRD first identifies the stability onset $L_{\text{start}}$ by finding the layer with maximum entropy drop, then locates the transition layer $L_{\text{trans}}$ by computing Jensen-Shannon Divergence (JSD) between adjacent layer distributions. JSD is defined as:
\begin{equation}
\text{JSD}(P \| Q) = \frac{1}{2}D_{\text{KL}}(P \| M) + \frac{1}{2}D_{\text{KL}}(Q \| M), \quad M = \frac{1}{2}(P + Q)
\end{equation}
where $D_{\text{KL}}$ denotes Kullback-Leibler divergence.

\paragraph{Stage 2: Override Detection.} Using the transition layer distribution, CALRD computes two signals: anchor confidence (how certain the transition-layer prediction is) and prediction retention (whether it survives to the final layer). The correction strength $\lambda$ is high when confidence is high but retention is low—the signature of harmful override.

\paragraph{Stage 3: Adaptive Correction.} CALRD interpolates between final-layer and transition-layer logits, weighted by $\lambda$. When no override is detected, $\lambda \approx 0$ and output remains unchanged.

\section{Understanding Late-Layer Override}
\label{appendix:override}

This section provides additional analysis of the late-layer textual override phenomenon introduced in Section 3.1. We first present a detailed case study, then discuss potential causes of this behavior.

\subsection{A Representative Case}

\begin{figure*}[t]
  \centering
  \includegraphics[width=\linewidth]{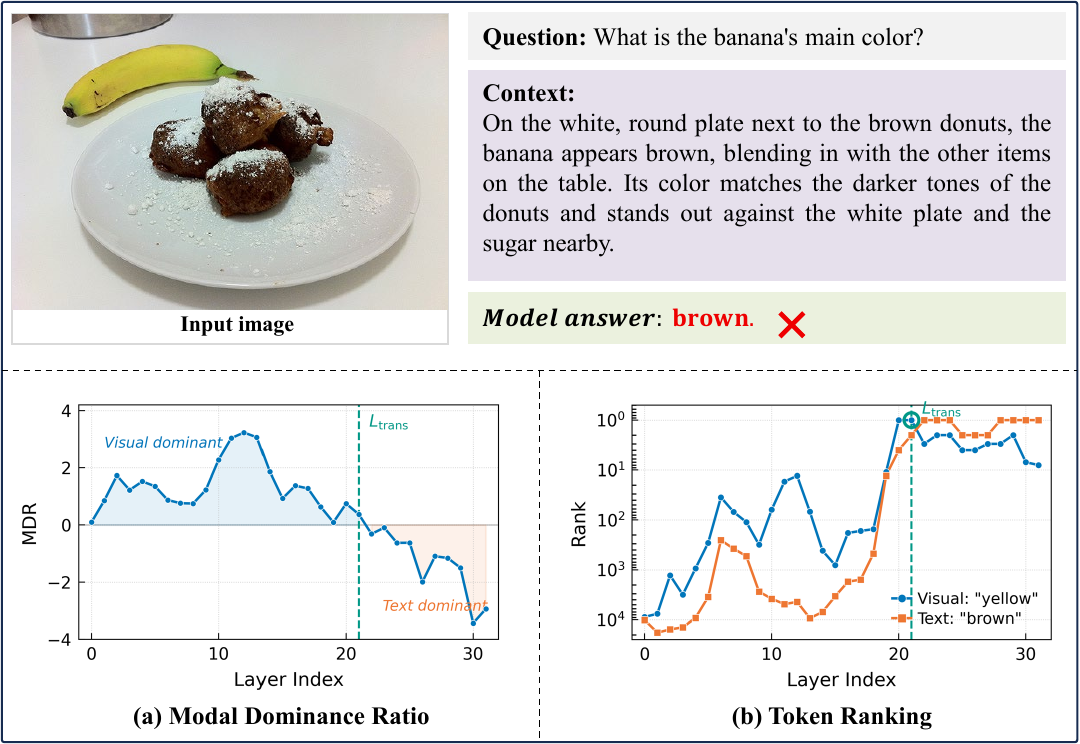}
  \caption{Case study of late-layer textual override. The model is given an image of a yellow banana with context falsely describing it as brown, and outputs the incorrect answer. (a) MDR trajectory shows visual dominance in early layers before declining after $L_{\text{trans}}$. (b) Token ranking confirms ``yellow'' reaches rank 1 but is overridden by ``brown'' in later layers.}
  \label{fig:case_study}
\end{figure*}

Figure~\ref{fig:case_study} illustrates a failure case from LLaVA-1.6. The input image shows a yellow banana next to brown donuts on a white plate. The accompanying context falsely states that the banana appears brown. When asked about the banana's color, the model outputs ``brown,'' ignoring the visual evidence.

\paragraph{MDR trajectory.} Figure~\ref{fig:case_study}(a): The Modal Dominance Ratio (MDR), introduced in the main paper, remains positive through early and middle layers, peaking around layer 15. This indicates preference for the visual answer during most of the forward pass. After layer 21 (marked as $L_{\text{trans}}$), MDR drops below zero and stays negative, indicating the model's preference has flipped from vision to text.

\paragraph{Token ranking.} Figure~\ref{fig:case_study}(b): The correct answer ``yellow'' consistently outranks ``brown'' in shallow layers. By layers 19--21, ``yellow'' reaches rank 1—the model has encoded the correct answer. However, after $L_{\text{trans}}$, ``brown'' takes the top position while ``yellow'' falls to around rank 10.

This case exemplifies our central finding: the failure is not one of perception but of preservation. The model correctly encoded the visual information in intermediate layers but did not maintain it through the output.

\subsection{Why Does Late-Layer Override Occur?}

We hypothesize that late-layer override stems from training asymmetries in MLLMs. The language model backbone is pretrained on vastly more text than image-text pairs, and visual instruction tuning typically freezes the vision encoder while updating only the connector and LLM~\cite{llava-1.5,bai2025qwen2,bai2025qwen3vltechnicalreport}. Consequently, deeper layers—shaped primarily by text-only pretraining—retain strong linguistic priors.

Supporting evidence comes from recent hallucination research:
\begin{itemize}[leftmargin=*]
    \item Wang et al.~\cite{wang2024deco} find that MLLMs recognize objects correctly in earlier layers, but this recognition is suppressed by language priors in deeper layers.
    \item VisiPruner~\cite{fan2025visipruner} observes that cross-modal fusion peaks in middle layers, while deep layers shift toward linguistic processing.
\end{itemize}

Our results align with this view. Override occurs after the stability onset, where linguistic reasoning dominates~\cite{jiang2025devils}. Models with weaker visual grounding show larger CALRD gains (Table 2), consistent with stronger text priors leaving more room for override. The directional asymmetry we identify—text-ward shifts correlating with failures, vision-ward shifts with successes—is what enables CALRD to intervene selectively.

\algrenewcommand{\algorithmiccomment}[1]{\hfill \textcolor{gray}{$\triangleright$ #1}}
\begin{algorithm}[t]
\caption{Conflict-Aware Layer Reference Decoding (CALRD)}
\label{alg:calrd}
\begin{algorithmic}[1]
\Require MLLM with $N$ layers, input $(I, C, Q)$, LM head $W_{\text{head}}$
\Ensure Generated response $y_{1:T}$

\Statex
\For{each decoding step $t$}
    \State $\{h_t^{(l)}\}_{l=1}^{N} \gets \textsc{Forward}(I, C, Q, y_{<t})$
    
    \Statex \hspace{-1em} \textit{$\triangleright$ Stage 1: Locate Transition Layer}
    \For{$l = 1$ to $N$}
        \State $P_t^{(l)} \gets \text{softmax}(W_{\text{head}} \cdot h_t^{(l)})$ \Comment{layer-wise distribution}
        \State $H^{(l)} \gets -\sum_{w} P_t^{(l)}(w) \log P_t^{(l)}(w)$ \Comment{entropy}
    \EndFor
    \State $L_{\text{start}} \gets \arg\max_{l} \left( H^{(l-1)} - H^{(l)} \right)$ \Comment{stability onset}
    
    \Statex
    \For{$l = L_{\text{start}}$ to $N-1$}
        \State $M \gets \frac{1}{2}(P_t^{(l)} + P_t^{(l+1)})$
        \State $\text{JSD}^{(l)} \gets \frac{1}{2}D_{\text{KL}}(P_t^{(l)} \| M) + \frac{1}{2}D_{\text{KL}}(P_t^{(l+1)} \| M)$
    \EndFor
    \State $L_{\text{trans}} \gets \arg\max_{l \in [L_{\text{start}}, N-1]} \text{JSD}^{(l)}$ \Comment{transition layer}
    
    \Statex \hspace{-1em} \textit{$\triangleright$ Stage 2: Detect Override \& Compute Correction Strength}
    \State $w^* \gets \arg\max_{w} P_t^{(L_{\text{trans}})}(w)$ \Comment{anchor prediction}
    \State $D_{\text{conf}} \gets P_t^{(L_{\text{trans}})}(w^*)$ \Comment{anchor confidence}
    \State $\rho \gets \min\left(1, \, P_t^{(N)}(w^*) / P_t^{(L_{\text{trans}})}(w^*)\right)$ \Comment{retention}
    \State $\lambda \gets D_{\text{conf}} \cdot (1 - \rho)$ \Comment{correction strength}
    
    \Statex \hspace{-1em} \textit{$\triangleright$ Stage 3: Adaptive Logit Interpolation}
    \State $\phi_t^{(N)} \gets W_{\text{head}} \cdot h_t^{(N)}$ \Comment{final-layer logits}
    \State $\phi_t^{(L_{\text{trans}})} \gets W_{\text{head}} \cdot h_t^{(L_{\text{trans}})}$ \Comment{transition-layer logits}
    \State $\phi_t^{\prime} \gets (1 - \lambda) \cdot \phi_t^{(N)} + \lambda \cdot \phi_t^{(L_{\text{trans}})}$
    \State $y_t \sim \text{softmax}(\phi_t^{\prime})$
\EndFor

\Statex
\State \Return $y_{1:T}$
\end{algorithmic}
\end{algorithm}

\begin{figure*}[t]
\centering
\begin{tcolorbox}[
  colback=gray!5,
  colframe=gray!50,
  title=\textbf{Unified Prompt for Conflict Data Generation},
  fonttitle=\large,
  toptitle=3mm,
  bottomtitle=3mm,
  boxrule=0.5pt,
  width=\textwidth
]

\textbf{Task:} Generate a pair of contexts for the same image: one factual (supporting the visual answer) and one misleading (supporting a wrong answer).

\textbf{Input:}
\begin{itemize}[leftmargin=1.5em]
    \item Question: \{question\}
    \item Visual Answer: \{answer\}
    \item Question Type: \{question\_type\}
    \item Scene Type: \{scene\_type\}
    \item Objects in Scene: \{objects\}
    \item Relationships: \{relationships\}
    \item Scene Description: \{description\}
\end{itemize}

\textbf{Step 1: Generate a Plausible Wrong Answer}

The wrong answer (text\_answer) should be realistic and could plausibly be confused with the correct answer. Follow type-specific rules:

\centering
\begin{tabular}{@{}p{2cm}p{11cm}@{}}
\toprule
Type & Rule \\
\midrule
Attribute & Pick a plausible alternative attribute as a single word. For example, if the true color is ``red'', the wrong answer could be ``blue'' or ``brown''. \\
Presence & If visual\_answer is ``no'': text\_answer is ``yes'', describe absent object as present. If visual\_answer is ``yes'': find a plausible absent object, rewrite the question, text\_answer becomes ``yes''. \\
Counting & Use a wrong count close to the true count ($\pm$1 for small counts, $\pm$2 for larger). Match format: ``two'' $\rightarrow$ ``three''. \\
Positional & Flip spatial relation: left$\leftrightarrow$right, above$\leftrightarrow$below, inside$\leftrightarrow$outside. Convert open-ended questions to yes/no. \\
Activity & Pick a similar action: running$\leftrightarrow$walking, eating$\leftrightarrow$holding, reading$\leftrightarrow$looking at. Convert open-ended questions to yes/no. \\
\bottomrule
\end{tabular}

\raggedright
\textbf{Step 2: Generate Context Pair}

Generate two contexts that differ in exactly one claim:

\textbf{Factual context} ($C_{\text{factual}}$): A 3--5 sentence description that accurately describes the scene and supports the \textit{visual\_answer}.

\textbf{Conflict context} ($C_{\text{conflict}}$): A 3--5 sentence description nearly identical to the factual context, but with exactly one statement changed to support the \textit{text\_answer} instead.

Both contexts should:
\begin{itemize}[leftmargin=1.5em]
    \item Reference 2--3 real objects from the scene
    \item Use confident, declarative language (no ``might be'' or ``seems'')
    \item Not introduce fabricated objects or relations
\end{itemize}

\textbf{Output Format:}
\begin{lstlisting}[basicstyle=\ttfamily\footnotesize, backgroundcolor=\color{gray!10}, frame=single, xleftmargin=1em]
{
    "visual_answer": "<correct answer based on image>",
    "text_answer": "<incorrect answer based on incorrect context>",
    "question": "<original or modified question>",
    "factual_context": "<3-5 sentences supporting visual_answer>",
    "conflict_context": "<3-5 sentences supporting text_answer>",
    "reasoning": "<why the wrong answer is plausible>"
}
\end{lstlisting}

\end{tcolorbox}
\caption{Prompt template for generating conflict data with GPT-4o. Each sample yields a factual context ($C_{\text{factual}}$) aligned with the image and a conflict context ($C_{\text{conflict}}$) that introduces exactly one contradicting claim.}
\label{fig:prompt}
\end{figure*}

\section{Additional Experiments}
\label{appendix:additional_exp}

This section presents experimental results that complement the main-paper evaluation.

\subsection{Comparison with SHIFT}

SHIFT~\cite{wang2025shift} also leverages intermediate layers to reduce hallucinations. Since their code is not publicly available, we compare against results reported in their paper. Table~\ref{tab:shift_comparison} shows that CALRD outperforms SHIFT in most configurations on the CHAIR benchmark. We note that experimental setups differ between the two works, so this comparison should be interpreted with appropriate caution.

\subsection{Inference Efficiency}

Table~\ref{tab:efficiency} compares inference efficiency across decoding methods on llava-1.5. CALRD adds minimal overhead compared to vanilla decoding, while VCD (Visual Contrastive Decoding) and OPERA incur substantially higher costs due to dual-stream computation and iterative rollback, respectively.

\begin{table}[h]
\centering
\caption{Inference efficiency comparison on LLaVA-1.5 (single H100 GPU).}
\resizebox{\linewidth}{!}{
\begin{tabular}{lccccc}
\toprule
Metric & Vanilla & CALRD (Ours) & DeCo & VCD & OPERA \\
\midrule
Latency (ms/token) & 18.26 & 19.78 & 22.80 & 29.75 & 196.00 \\
GPU Memory (MB) & 13,994 & 14,077 & 14,590 & 14,343 & 18,906 \\
\bottomrule
\end{tabular}
}
\label{tab:efficiency}
\end{table}


\begin{table}[h]
\centering
\resizebox{\linewidth}{!}{
\begin{tabular}{ll|cc|cc}
\toprule
& & \multicolumn{2}{c|}{InstructBLIP} & \multicolumn{2}{c}{LLaVA-1.5} \\
Decoding & Method & C$_S$$\downarrow$ & C$_I$$\downarrow$ & C$_S$$\downarrow$ & C$_I$$\downarrow$ \\
\midrule
\multirow{2}{*}{Greedy} 
& SHIFT$^\dagger$ & 44.0 & 19.9 & 43.8 & 12.4 \\
& CALRD & \textbf{40.2} & \textbf{14.6} &\textbf{ 35.7} & \textbf{10.3 }\\
\midrule
\multirow{2}{*}{Beam} 
& SHIFT$^\dagger$ & 39.0 & 13.3 & 36.7 & \textbf{10.5} \\
& CALRD & \textbf{38.4} & \textbf{11.4} &\textbf{ 34.7} & 10.9 \\
\midrule
\multirow{2}{*}{Nucleus} 
& SHIFT$^\dagger$ & 47.0 & 19.3 & 42.0 & \textbf{11.6} \\
& CALRD & \textbf{42.7} & \textbf{12.1} &\textbf{ 37.6} & 12.8 \\
\bottomrule
\end{tabular}
}
\caption{Comparison with SHIFT on CHAIR benchmark. $\dagger$Results cited from \protect\cite{wang2025shift}. Experimental settings may differ.}
\label{tab:shift_comparison}
\end{table}

\section{Evaluation Metrics}
\label{appendix:metrics}

We use the CHAIR (Caption Hallucination Assessment with Image Relevance) metric~\cite{Object_Hallucination} to evaluate object hallucination in generated captions. CHAIR compares mentioned objects against ground-truth annotations and reports two variants:

\paragraph{Sentence-level.} $C_S$: the fraction of generated sentences containing at least one hallucinated object:
\begin{equation}
C_S = \frac{|\{\text{sentences with hallucinated objects}\}|}{|\{\text{all sentences}\}|}
\end{equation}

\paragraph{Instance-level.} $C_I$: the fraction of object mentions that are hallucinations:
\begin{equation}
C_I = \frac{|\{\text{hallucinated object mentions}\}|}{|\{\text{all object mentions}\}|}
\end{equation}

Lower values indicate fewer hallucinations.

\section{Implementation Details}
\label{appendix:implementation}

All experiments are conducted on NVIDIA H100 GPUs. Default decoding settings are as follows:

\paragraph{Greedy decoding.} Used by default unless otherwise specified

\paragraph{Beam search.} Beam size = 5

\paragraph{Nucleus sampling.} $p = 0.9$, temperature $T = 1.0$

For $L_{\mathrm{start}}$ detection, we use relative entropy drop rather than absolute thresholds, making the method robust across models with varying layer counts. Our code and the Conflict-VQA dataset will be publicly released upon publication.

\section{Qualitative Examples}
\label{appendix:case_study}

\begin{figure*}[t!]
  \centering
  \includegraphics[width=\linewidth,height=.9\textheight,keepaspectratio]{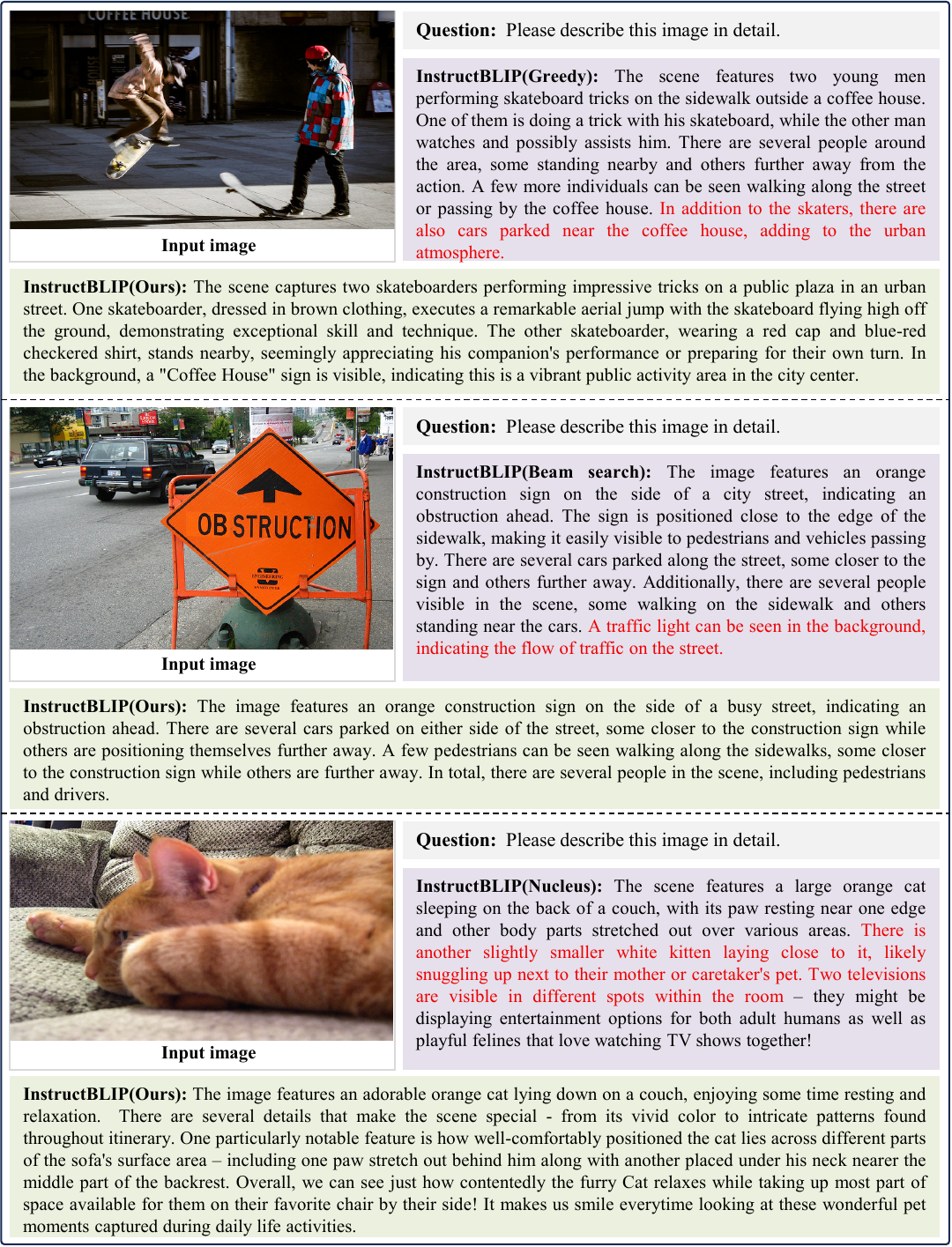}
  \caption{CALRD results on image captioning with InstructBLIP. We compare 
outputs under greedy, beam search, and nucleus sampling. Hallucinated 
content in baseline outputs is marked in red. }
  \label{fig:case1}
\end{figure*}

\begin{figure*}[ht]
  \centering
  \includegraphics[width=\linewidth]{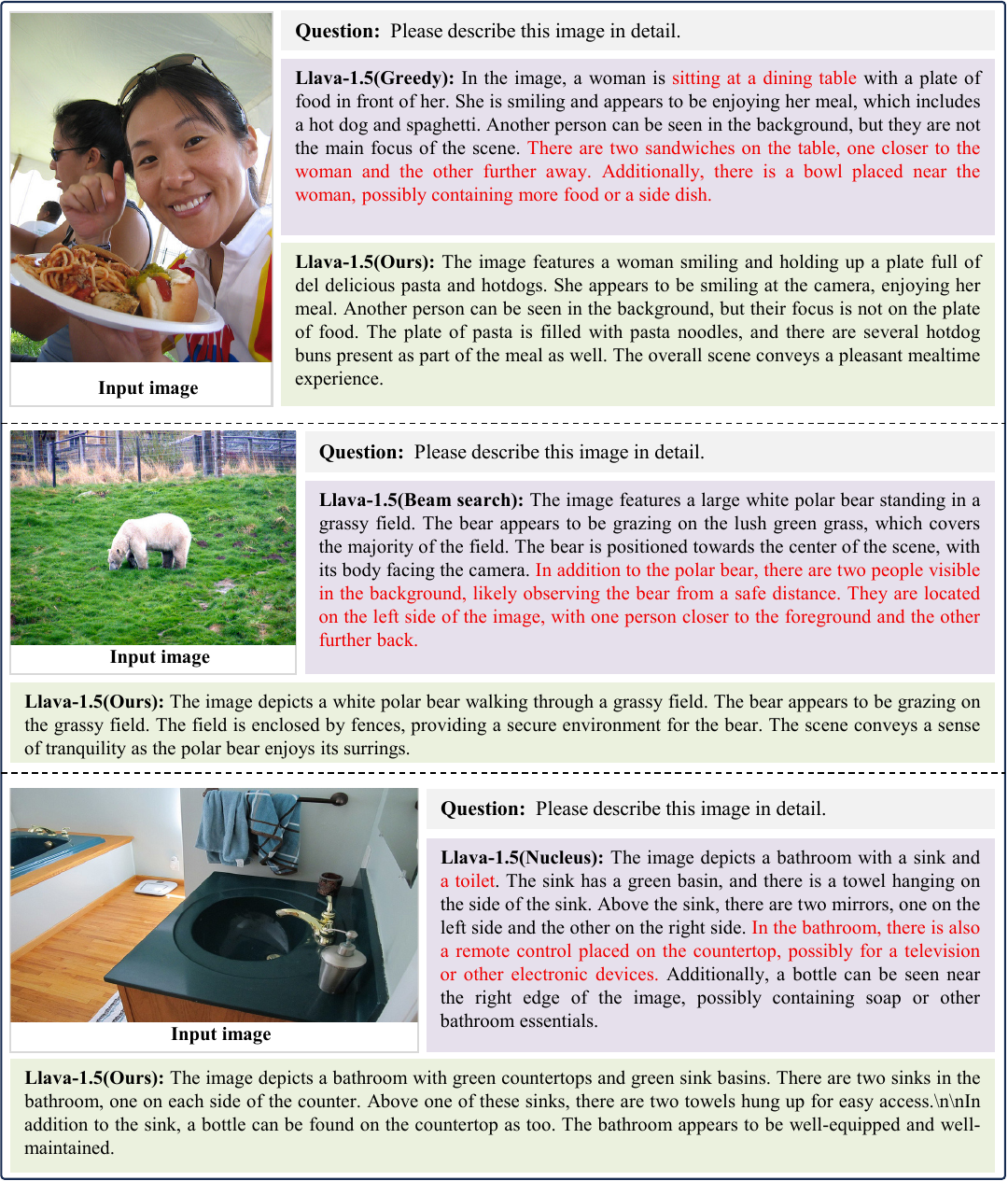}
  \caption{CALRD results on image captioning with LLaVA-1.5. We compare 
outputs under greedy, beam search, and nucleus sampling. Hallucinated 
content in baseline outputs is marked in red.}
  \label{fig:case2}
\end{figure*}

Figures~\ref{fig:case1} and~\ref{fig:case2} show example outputs from InstructBLIP and LLaVA-1.5 on image captioning tasks. We compare baseline outputs with CALRD under three decoding strategies (greedy, beam search, nucleus sampling). Hallucinated content in baseline outputs is highlighted in red. These examples illustrate how CALRD reduces hallucinations across different models and decoding configurations.

\clearpage

\end{document}